\documentclass{article}

\PassOptionsToPackage{numbers, sort&compress}{natbib}

    \usepackage[final]{neurips_2024}

\usepackage{multicol, multirow}
\usepackage[utf8]{inputenc} 
\usepackage[T1]{fontenc}    
\usepackage{hyperref}       
\usepackage{url}            
\usepackage{booktabs}       
\usepackage{amsfonts}       
\usepackage{nicefrac}       
\usepackage{microtype}      
\usepackage{xcolor}         
\usepackage{amsmath}
\usepackage{graphicx}
\usepackage{arydshln}
\usepackage{float}
\usepackage{wrapfig}
\usepackage{subcaption}

\title{PiLaMIM: Toward Richer Visual \\ Representations by Integrating Pixel and Latent Masked Image Modeling}

\author{
  Junmyeong Lee \And
  Eui Jun Hwang \And
  Sukmin Cho \And
  Jong C. Park \And \\[-6ex]
  School of Computing \\
  Korea Advanced Institute of Science and Technology\\
  \texttt{\{david516, ehwa20, nelllpic, jongpark\}@kaist.ac.kr} \\
}

\begin{document}

\maketitle

\begin{abstract}
In Masked Image Modeling (MIM), two primary methods exist: Pixel MIM and Latent MIM, each utilizing different reconstruction targets, raw pixels and latent representations, respectively. Pixel MIM tends to capture low-level visual details such as color and texture, while Latent MIM focuses on high-level semantics of an object. However, these distinct strengths of each method can lead to suboptimal performance in tasks that rely on a particular level of visual features. To address this limitation, we propose PiLaMIM, a unified framework that combines Pixel MIM and Latent MIM to integrate their complementary strengths. Our method uses a single encoder along with two distinct decoders: one for predicting pixel values and another for latent representations, ensuring the capture of both high-level and low-level visual features. We further integrate the \(\texttt{[CLS]}\) token into the reconstruction process to aggregate global context, enabling the model to capture more semantic information. Extensive experiments demonstrate that PiLaMIM outperforms key baselines such as MAE, I-JEPA and BootMAE in most cases, proving its effectiveness in extracting richer visual representations. The code is available at \href{https://github.com/joonmy/PiLaMIM.git}{https://github.com/joonmy/PiLaMIM.git}.

\end{abstract}

\section{Introduction}
Masked Image Modeling (MIM) has shown to be an effective pre-training method for Vision Transformers (ViT)~\cite{DBLP:ViT}, achieving notable successes in various downstream tasks. MIM involves masking a significant portion of an image and training the model to predict the masked regions based solely on the visible context, using various reconstruction targets such as pixels~\cite{DBLP:MAE, DBLP:simMIM,DBLP:videomae, DBLP:convmae}, latent representations~\cite{DBLP:iBOT,DBLP:I-JEPA,DBLP:data2vec, DBLP:CAE}, discrete visual tokens~\cite{DBLP:BeiT,DBLP:beitv2}, and hand-crafted features~\cite{DBLP:MaskFeat}. 
The reconstruction target in pre-training plays a crucial role in determining representation quality, directly aligned with high performance of certain downstream tasks~\cite{DBLP:latentMIM,DBLP:I-JEPA,DBLP:MaskFeat, DBLP:beitv2,DBLP:BeiT,DBLP:pixMIM}.

In particular, when pre-trained with pixels (\textbf{Pixel MIM}) and latent representations (\textbf{Latent MIM}) as the reconstruction targets, the visual representations learned by the two methods tend to have contrasting characteristics~\cite{DBLP:latentMIM,DBLP:I-JEPA}. Pixel MIM focuses on low-level details such as color, edges and texture, making it particularly effective for tasks that require fine-grained visual information such as object counting and depth prediction. By contrast, Latent MIM captures high-level semantics, excelling at tasks such as image classification, where abstract and semantic understanding of an object is more important. 

However, those specialized strengths of each MIM method can also be viewed as obvious weaknesses: (i) Pixel MIM, with its emphasis on low-level details, is limited to capture high-level semantics. This results in lower performance on tasks that require deeper semantic understanding, especially without fine-tuning~\cite{DBLP:MAE, DBLP:pixMIM, DBLP:latentMIM, DBLP:I-JEPA}. (ii) Latent MIM, on the other hand, uses latent representations as learning targets, where pixel details are potentially removed. As a result, it tends to perform worse than Pixel MIM in tasks that rely on low-level details~\cite{DBLP:latentMIM, DBLP:I-JEPA}. 
These limitations indicate that neither Pixel MIM nor Latent MIM is entirely optimal on its own, but their strengths may complement each other's weaknesses in understanding the visual context.
This presents an opportunity to achieve richer visual representations by combining both reconstruction targets in pre-training ViT~\cite{DBLP:ViT} with MIM.
 
Building on these observations, we integrate both \textbf{Pi}xel MIM and \textbf{La}tent MIM methods into a unified framework, \textbf{PiLaMIM}, which complements each method and enables richer visual representations.
This structure employs a shared context encoder along with two distinct decoders, each serving a different purpose: one designed to reconstruct pixel values, and the other focused on predicting latent representations. 
Moreover, we enhance the prediction of latent representations by incorporating the \(\texttt{[CLS]}\) token in the process, following iBOT~\cite{DBLP:iBOT}.
The inclusion of the \(\texttt{[CLS]}\) token is critical, as it aggregates global context across the entire image, encouraging the model to go beyond local patches and capture more abstract, semantically rich information~\cite{DBLP:iBOT,DBLP:beitv2}.

To highlight the importance of capturing both high-level and low-level visual features, we design our experiments in two distinct categories: (i) image classification tasks that rely on high-level semantics using ImageNet-1K~\cite{DBLP:imagenet}, CIFAR10~\cite{cifar}, CIFAR100~\cite{cifar}, iNaturalist2021~\cite{DBLP:iNat}, and Places365~\cite{DBLP:places}, and (ii) object counting and depth prediction tasks that depend on low-level details using Clevr/Count~\cite{DBLP:Clevr} and Clevr/Dist~\cite{DBLP:Clevr}. 
We experimentally show that PiLaMIM consistently outperforms MAE~\cite{DBLP:MAE} and I-JEPA~\cite{DBLP:I-JEPA}, which are leading methods in Pixel MIM and Latent MIM, respectively. Moreover, PiLaMIM shows better performance than BootMAE~\cite{DBLP:bootmae}, which similarly attempts at integrating both MIM methods. These results demonstrate that our method excels at extracting richer visual representation by effectively capturing both high-level and low-level visual features. Furthermore, we find that simply utilizing the \(\texttt{[CLS]}\) token in the reconstruction process can significantly enhance the performance in both high-level and low-level visual tasks. 


\section{Method}
\begin{figure}[t!]
    \centering    \includegraphics[width=.95\textwidth]{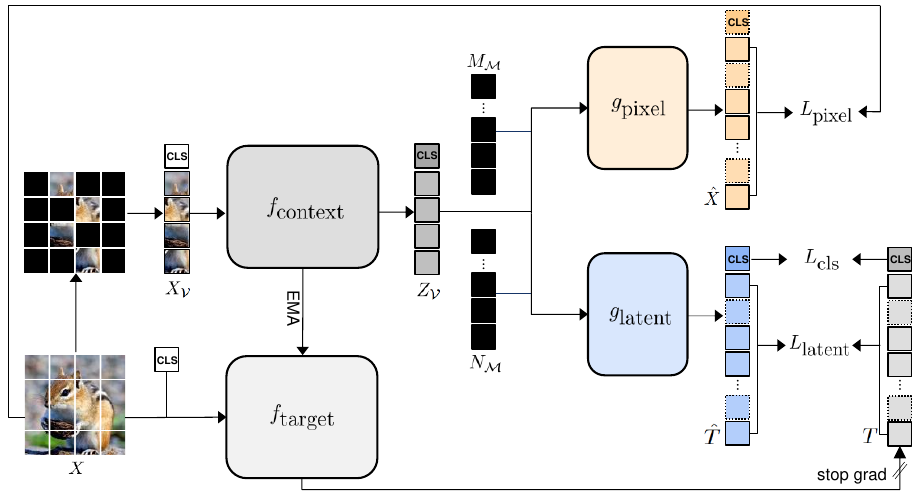}
    \caption{\small
    An overview of the PiLaMIM framework. The context encoder \(f_{\textnormal{context}}\) processes the visible patches \(X_{\mathcal{V}}\) to produce encoded tokens \(Z_{\mathcal{V}}\). These tokens are then fed into two decoders: the pixel decoder \(g_{\textnormal{pixel}}\) and \(g_{\textnormal{latent}}\), resulting in predicted pixel values \(\hat{X}\) and latent representations \(\hat{T}\), respectively. The target encoder \(f_{\textnormal{target}}\) processes the patched image \(X\) to provide target latent representation \(T\). For training, we define three loss functions: \(L_{\textnormal{pixel}}\) and \(L_{\textnormal{latent}}\) for masked patches, and \(L_{\textnormal{cls}}\) for the \(\texttt{[CLS]}\) token. 
    }
    \label{fig:maejepa}
    \vspace{-1.em}
\end{figure}

To capture both high-level and low-level visual features, as illustrated in Figure~\ref{fig:maejepa}, we extend the Masked Autoencoder (MAE)~\cite{DBLP:MAE} by integrating two distinct decoders: a pixel decoder and a latent decoder. The pixel decoder is responsible for reconstructing at the pixel level, while the latent decoder focuses on reconstructing latent level features. The specifics of each component in our method are described in the following sections.

\paragraph{Context Encoder.}
The input image \(I\) is initially divided into \(N\) non-overlapping patches \(X = \{x_{i}\}_{i=1}^N\), where \(x_{i}\) denotes the $i$-th patch of the image. A portion \(k\) of these patches is randomly selected to be masked, resulting in two sets: visible patches \(X_{\mathcal{V}} = \{x_{i}\}_{i \in \mathcal{V}}\) and masked patches \(X_{\mathcal{M}} = \{x_{i}\}_{i \in \mathcal{M}}\), where \(\mathcal{V}\) and \(\mathcal{M}\) represent the index sets of the visible and masked patches, respectively.
The context encoder  \(f_{\textnormal{context}}\) processes only the visible patches, which are first linearly projected to match the encoder's dimensions. Then, positional embeddings \(P_{\mathcal{V}}\) are added along with the \(\texttt{[CLS]}\) token prepended. These processed inputs are subsequently fed into the ViT~\cite{DBLP:ViT} layers to output the latent representation \(Z_{\mathcal{V}}\):
\begin{equation}
    Z_{\mathcal{V}} = f_{\textnormal{context}}(\texttt{[CLS]}, X_{\mathcal{V}}, P_{\mathcal{V}})
\end{equation}

\paragraph{Target Encoder.}
The target encoder \(f_{\textnormal{target}}\) shares the same structure as the context encoder and is initialized with the same weights. 
It is updated using the exponential moving average (EMA) of the context encoder weights~\cite{DBLP:dino}, following the update rule: \(\theta_{\textnormal{target}}^{(t)} = \lambda \theta_{\textnormal{target}}^{(t-1)} + (1 - \lambda) \theta_{\textnormal{context}}^{(t)}\), where \(\lambda\) is the momentum value and \(\theta\) indicates model parameters. The target encoder \(f_{\textnormal{target}}\) takes the entire patched image \(X\) along with positional embedding \(P\)  and the \(\texttt{[CLS]}\) token as input to extract the target latent representation \(T = \{t_{i}\}_{i=0}^N\):
\begin{equation}
    T = f_{\textnormal{target}}(\texttt{[CLS]}, X, P)
\end{equation}

\paragraph{Pixel and Latent Decoder.}
The decoder consists of two components: pixel decoder \(g_{\textnormal{pixel}}\) for pixel reconstruction and latent decoder \(g_{\textnormal{latent}}\) for latent representation reconstruction. 
After the encoding process, the encoded tokens are fed into each decoder and linearly projected to match the decoder's dimensions. These tokens and the learnable tokens \(M_{\mathcal{M}}\) and \(N_{\mathcal{M}}\), which are introduced to predict the masked patches \(X_{\mathcal{M}}\), are rearranged back to their original position, after which positional embeddings \(P\) are added. This is then passed through ViT~\cite{DBLP:ViT} layers and projected again to match each target's dimensions, generating the predicted pixel values \(\hat{X}\) and latent representations \(\hat{T}\):
\begin{equation}
\hat{X} = g_{\textnormal{pixel}}(Z_{\mathcal{V}}, M_{\mathcal{M}}, P), \quad
\hat{T} = g_{\textnormal{latent}}(Z_{\mathcal{V}}, N_{\mathcal{M}}, P)
\end{equation}

\paragraph{Training Objective.}
The outputs from each decoder are used to define the objective function using Mean Squared Error (MSE) between the pixel values and latent representations extracted by the target encoder. We calculate only masked patches, following MAE~\cite{DBLP:MAE}. Additionally, for latent representations, we include calculations for the \(\texttt{[CLS]}\) token along with the masked patches.
The objective functions for pixel and latent decoders are respectively computed as follows:
\begin{equation}
    \textstyle{
    L_{\textnormal{pixel}} = \frac{1}{D_x\cdot|\mathcal{M}|}\sum\limits_{i \in \mathcal{M}}\|\hat{x}_{i}-x_{i}\|^2_2, \quad
    L_{\textnormal{latent}} = \frac{1}{D_t\cdot|\mathcal{M}|}\sum\limits_{i \in \mathcal{M}}\|\hat{t}_{i}-t_{i}\|^2_2, \quad
    L_{\textnormal{cls}} = \frac{1}{D_t}\|\hat{t}_{0} - t_{0}\|^2_2
    }
\end{equation}
Here, \(D_x\) and \(D_t\) are the dimensions of \(x_i\) and \(t_i\), respectively.
The overall objective function for training is defined as a sum of \(L_{\textnormal{pixel}}\), \(L_{\textnormal{latent}}\) and \(L_{\textnormal{cls}}\):
\begin{equation}
L = L_{\textnormal{pixel}} + L_{\textnormal{latent}} + L_{\textnormal{cls}}
\end{equation}

\section{Experiments}
\paragraph{Dataset.} 
To validate the effectiveness of our method, we conducted experiments in two categories, following~\cite{DBLP:I-JEPA}:
(i) Image classification task which requires high-level semantics using ImageNet-1K, CIFAR10, CIFAR100, iNaturalist2021 and Places365 datasets. 
(ii) Object counting and depth prediction tasks which require low-level details using Clevr/Count and Clevr/Dist datasets. 
For iNaturalist2021, a mini version was used considering an appropriate data size. Since we performed pre-training on ImageNet-1K before conducting downstream tasks, we will refer to all datasets except ImageNet-1K as out-of-domain data.

\paragraph{Implementation Details.} We pretrained the baselines and our method on ImageNet-1K and evaluated their performance through linear probing using the ViT-Base architecture. We compare our method against MAE~\cite{DBLP:MAE} and I-JEPA~\cite{DBLP:I-JEPA}, which are prominent methods for Pixel MIM and Latent MIM, respectively, and BootMAE~\cite{DBLP:bootmae}, which combines these two methods. I-JEPA was pretrained for 600 epochs, MAE and BootMAE for 800 epochs, and PiLaMIM for both 600 and 800 epochs.
During linear probing, both I-JEPA and BootMAE used the average pooled patch token from the last layer of the encoder, following the settings of each paper.
For MAE and PiLaMIM, we used the \(\texttt{[CLS]}\) token from the same layer. However, for Clevr/Dist dataset, we observed a large performance gap between the \(\texttt{[CLS]}\) token and the average pooled patch token, so we standardized the evaluation using the \(\texttt{[CLS]}\) token. These tokens were processed through a linear layer, preceded by batch normalization, and used for final performance evaluation. All linear probing was performed over 100 epochs, with the best performance reported afterwards. Detailed model architectures and training settings are described in \ref{sec:appen_architecture} and \ref{sec:appen_training}.


\paragraph{Main Results}
Table~\ref{tab:performance-comparison} (a) shows the performance of the high-level task with image classification evaluated using top-1 accuracy. 
As previously known~\cite{DBLP:I-JEPA}, I-JEPA, which targets latent representations, outperforms MAE on most datasets. However, PiLaMIM significantly surpasses I-JEPA across all datasets except for ImageNet-1K, even at 600 epochs, particularly excelling on out-of-domain data. Specifically, our method achieves 92.4\% accuracy on CIFAR10 and 74.2\% on CIFAR100, which are 6.1\% and 8.1\% higher than I-JEPA, respectively. Furthermore, PiLaMIM outperforms BootMAE across most datasets, indicating that the integration of the \texttt{[CLS]} token enhances the model's ability to capture high-level semantics. Meanwhile, Table~\ref{tab:performance-comparison} (b) presents the performance of the low-level tasks such as object counting and depth prediction.
As in prior studies~\cite{DBLP:I-JEPA}, MAE, which targets raw pixels, outperforms I-JEPA in both tasks. Notably, PiLaMIM outperforms MAE across all tasks, achieving a 7.2\% increase to 83.8\% on Clevr/Count and a 2.8\% increase to 67.8\% on Clevr/Dist. Moreover, PiLaMIM performs comparably to or slightly better than BootMAE, particularly achieving a higher accuracy on Clevr/Count, demonstrating its effectiveness in capturing low-level details. 
We performed an additional visual analysis using t-SNE~\cite{t-sne} in \ref{sec:appen_vis_anal}.

\begin{table}[t!]
\centering
\caption{\small Performance comparison on high-level and low-level visual tasks. The highest performances are highlighted in bold.}
\label{tab:performance-comparison}
\resizebox{.99\textwidth}{!}{
\begin{tabular}{l c ccccc cc}
\toprule
&  & \multicolumn{5}{c}{(a) High-level Task} & \multicolumn{2}{c}{(b) Low-level Task} \\ \cmidrule(l{2pt}r{2pt}){3-7} \cmidrule(l{2pt}r{2pt}){8-9}
 \textbf{Method} & \textbf{Epochs} & ImageNet-1K & CIFAR10 & CIFAR100 & iNat2021 & Places365 & Clevr/Count & Clevr/Dist \\ 
\midrule
\textbf{MAE}~\cite{DBLP:MAE} & 800 & 61.1 & 86.7 & 66.0 & 25.8 & 44.2 & 76.6 & 65.0 \\ 
\textbf{I-JEPA}~\cite{DBLP:I-JEPA} & 600 & 67.7 & 86.3 & 66.1 & 27.7 & 45.3 & 71.6 & 62.0 \\ 
\textbf{BootMAE}~\cite{DBLP:bootmae} & 800 & 67.6 & 90.8 & 72.7 & 26.6 & \textbf{48.8} & 82.6 & \textbf{68.7}\\ 
\noalign{\vskip 0.3ex}\cdashline{1-9}\noalign{\vskip 0.7ex}
\multirow{2}{*}{\textbf{PiLaMIM}} & 600 & 67.0 & 91.0 & 73.0 & 29.5 & 47.0 & 80.7 & 66.8 \\ 
 & 800 & \textbf{69.2} & \textbf{92.4} &\textbf{74.2} & \textbf{31.0} & 47.6 & \textbf{83.8} & 67.8 \\ 
\bottomrule
\end{tabular}
}
\vspace{-1.em}
\end{table}

These results demonstrates that our method, which targets pixels and latent representations, effectively captures both high-level and low-level visual features. Furthermore, the significant performance improvements observed in both categories of experiments indicate the importance of integrating both levels of visual features, which complement each other and lead to a richer visual representation. 



\begin{wraptable}{R}{.4\textwidth}

\vspace{-1.em}
\caption{\small Ablation study for \(\texttt{[CLS]}\) token.}
\vspace{-.2em}
\label{tab:ablation}
\resizebox{.4\textwidth}{!}{
\begin{tabular}{l c c}
\toprule
 \textbf{Method} & CIFAR100 & Clevr/Count \\
\midrule
\textbf{Ours w/o \(\texttt{[CLS]}\)} & 72.3 & 82.6 \\
\textbf{Ours w \(\texttt{[CLS]}\)} & \textbf{74.2} & \textbf{83.8} \\
\bottomrule
\end{tabular}
}

\end{wraptable}

\paragraph{Ablation.}
We conducted an ablation study to explore the importance of the \(\texttt{[CLS]}\) token. Table~\ref{tab:ablation} shows the linear probing performance of PiLaMIM, pre-trained for 800 epochs with and without the \(\texttt{[CLS]}\) token. Even without the \(\texttt{[CLS]}\) token, PiLaMIM still outperforms MAE and I-JEPA on CIFAR100 and Clevr/Count. However, when the \(\texttt{[CLS]}\) token is included, the performance improves by 1.9\% on CIFAR100 and 1.2\% on Clevr/Count. This indicates that simply utilizing the \texttt{[CLS]} token can boost performance in both high-level and low-level visual tasks, suggesting potential for further improvements with its expanded use.

\section{Conclusion}
We proposed PiLaMIM, a unified framework that combines Pixel MIM and Latent MIM, to effectively capture both high-level and low-level visual features. Our experiments showed that PiLaMIM consistently outperformed existing methods across high-level and low-level visual tasks, demonstrating the benefits of integrating both types of visual features. Notably, the inclusion of the \(\texttt{[CLS]}\) token further improved the performance in both visual tasks, highlighting its importance in capturing both levels of visual features more effectively. These findings demonstrate that PiLaMIM is capable of extracting rich and robust visual representations, making it well-suited for a wide range of visual tasks.

\section{Acknowledgements}
This work was supported by the Institute for Information and communications Technology Promotion (IITP) grant funded by the Korea government (MSIT) (No. 2022-0-00010, Development of Korean sign language translation service technology for the deaf in medical environment), and the Artificial intelligence industrial convergence cluster development project funded by the Ministry of Science and ICT (MSIT, Korea) \& Gwangju Metropolitan City.

\bibliographystyle{abbrvnat}
\bibliography{reference}







\appendix

\section{Appendix}

\subsection{Limitations}
Due to limited resources, we were unable to consider a broader range of visual tasks such as semantic segmentation and object detection. Additionally, we could only pre-train our method for up to 800 epochs, and were unable to explore the potential performance improvements with extended training. While MAE~\cite{DBLP:MAE}, I-JEPA~\cite{DBLP:I-JEPA}, and PiLaMIM were all trained in our environment, BootMAE\footnote{\url{https://github.com/LightDXY/BootMAE}}~\cite{DBLP:bootmae} was used with a publicly available checkpoint for linear probing due to time constraints.

\subsection{Visual Analysis}\label{sec:appen_vis_anal}
\begin{figure}[ht]
    \centering
    \begin{subfigure}[b]{0.3\textwidth}
        \includegraphics[width=\textwidth]{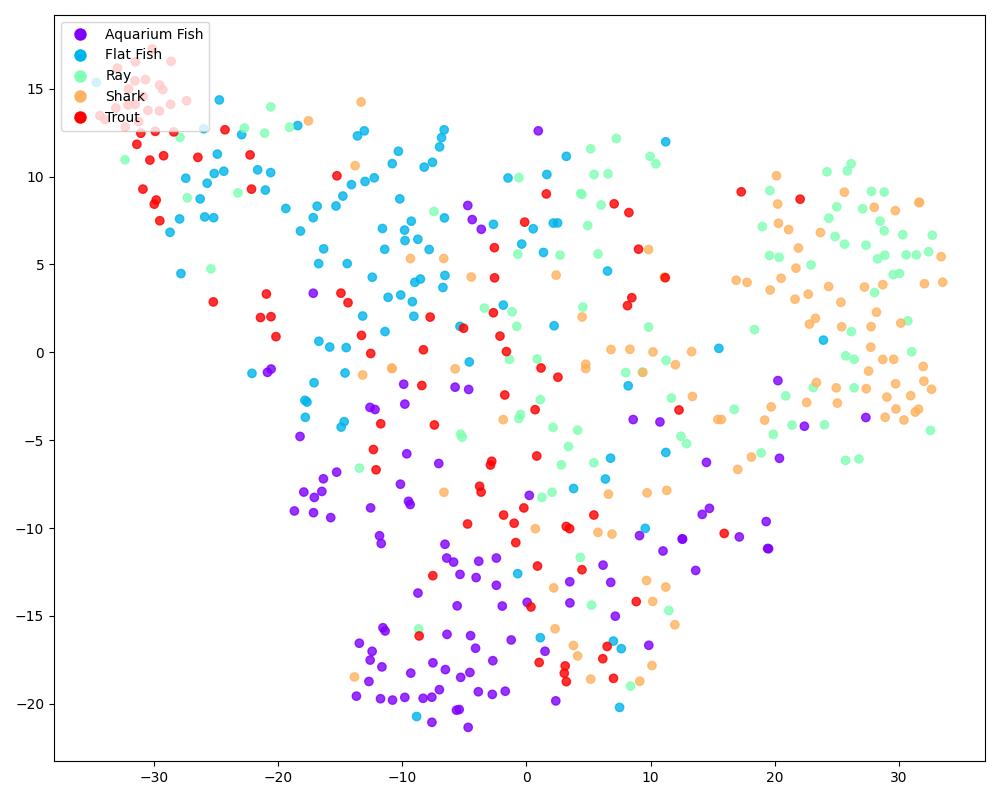}
        \caption{MAE}
        \label{fig:mae}
    \end{subfigure}
    \hfill
    \begin{subfigure}[b]{0.3\textwidth}
        \includegraphics[width=\textwidth]{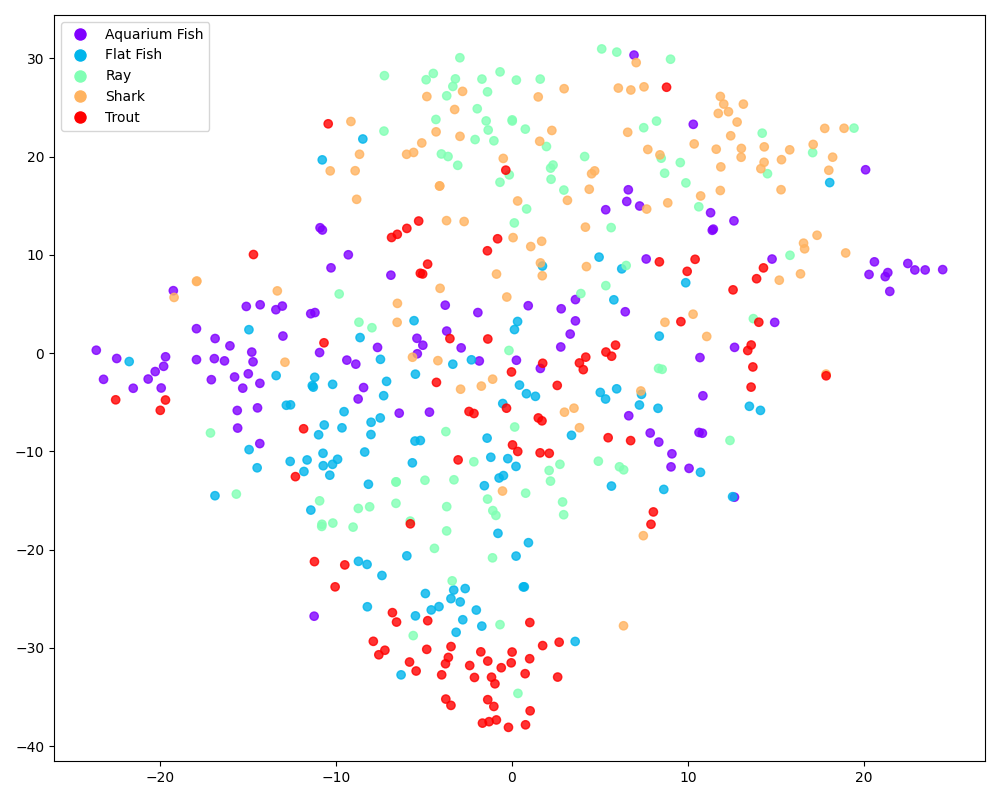}
        \caption{I-JEPA}
        \label{fig:jepa}
    \end{subfigure}
    \hfill
    \begin{subfigure}[b]{0.3\textwidth}
        \includegraphics[width=\textwidth]{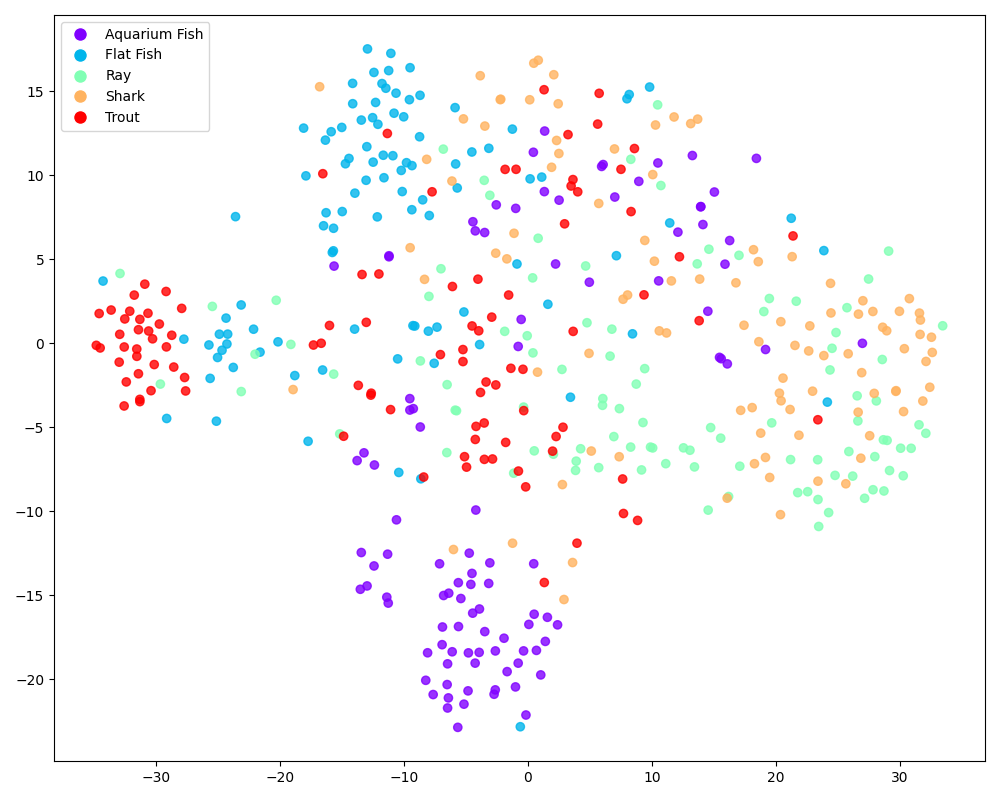}
        \caption{PiLaMIM}
        \label{fig:PiLaMIM}
    \end{subfigure}
    \caption{The t-SNE visualizations of CIFAR100 ``Fish'' superclass images. Different colors represent different subclasses. They are better viewed by zooming in.}
    \label{fig:tsne-comparisons}
\end{figure}

Figure \ref{fig:tsne-comparisons} illustrates the t-SNE visualizations of images from the ``Fish'' superclass in the CIFAR100 validation set, using the encoders trained with MAE, I-JEPA, and PiLaMIM on ImageNet-1K. CIFAR100 is comprised of 20 superclasses, each containing 5 subclasses. Images within the same superclass share highly similar semantics, which increases the difficulty of accurate classification. Compared to MAE and I-JEPA, which target either pixel or latent representations, PiLaMIM shows better subclass clustering. This demonstrates that targeting both pixel and latent representation in MIM extracts richer and more robust visual representations. 

\subsection{RankMe}
Table \ref{tab:rankme} shows the RankMe~\cite{DBLP:rankme} scores for each model on the CIFAR100 dataset. RankMe is a metric that allows for the measurement of representational quality without performing downstream tasks. PiLaMIM achieved significantly higher RankMe scores compared to MAE and I-JEPA, indicating that the quality of representations extracted by PiLaMIM is superior.
\begin{table}[ht]
\centering

\caption{RankMe scores on CIFAR100.}

\label{tab:rankme}
\begin{tabular}{l c}
\toprule
  Method & RankMe \raisebox{0.1em}{$\uparrow$} \\
\midrule
MAE & 217.80 \\
I-JEPA & 226.58 \\
PiLaMIM & \textbf{254.78} \\
\bottomrule
\end{tabular}
\end{table}


\subsection{Architecture details}\label{sec:appen_architecture}
The encoder for all models consists of 12 ViT~\cite{DBLP:ViT} blocks with 768 dimensions and 12 heads. The decoder configurations vary across models: MAE has 8 ViT blocks with 512 dimensions and 16 heads; I-JEPA and PiLaMIM have 6 ViT blocks with 384 dimensions and 12 heads; BootMAE has 2 cross-attention based ViT blocks with 512 dimensions and 12 heads.

The input image's height and width are 224, with a patch size of 16, resulting in a total of 196 patches being fed into the model. The masking ratio is 0.75, and no normalization was applied to the target pixels and latent representations. The \(\lambda\) used for updating the target encoder was initialized at 0.996 and linearly increased to 1.0 throughout pre-training.

\subsection{Training details}\label{sec:appen_training}
Table \ref{tab:pretrain} shows the pre-training settings on ImageNet-1K, while Table \ref{tab:linear} presents the linear probing settings. For linear probing, we used a batch size of 16384 for ImageNet-1K~\cite{DBLP:imagenet} and Places365~\cite{DBLP:places}, 8192 for iNat2021~\cite{DBLP:iNat}, and 1024 for CIFAR10~\cite{cifar}, CIFAR100~\cite{cifar}, Clevr/Count~\cite{DBLP:Clevr}, and Clevr/Dist~\cite{DBLP:Clevr}. We pre-trained PiLaMIM for 800 epochs using four NVIDIA A100 80GB GPUs over approximately 120 hours. Each linear probing downstream task was completed within 24 hours, using a single GPU for CIFAR10, CIFAR100, Clevr/Count, and Clevr/Dist, while 4 GPUs were used for the other datasets.



\begin{table}[h]
    \centering
    \caption{Pre-training setting on ImageNet-1K.}
    \begin{tabular}{l | c}
    \hline
    config & value \\
    \hline
    optimizer   & AdamW~\cite{DBLP:adamw} \\
    base learning rate & 1.5e-4 \\
    weight decay & 0.05 \\
    optimizer momentum & \(\beta_{1}\) = 0.9, \(\beta_{2}\) = 0.95 \\
    batch size & 2048 \\
    learning rate schedule & cosine decay \\
    warmup epochs & 40 \\
    training epochs & 800 \\
    augmentation & RandomResizedCrop \\
    \hline
    \end{tabular}
    \label{tab:pretrain}
\end{table}

\begin{table}[h]
    \centering
    \caption{Linear probing setting.}
    \begin{tabular}{l | c}
    \hline
    config & value \\
    \hline
    optimizer   & LARS~\cite{LARS} \\
    base learning rate & 3 \\
    weight decay & 0 \\
    optimizer momentum & 0.9 \\
    batch size &  varies by dataset \\
    learning rate schedule & cosine decay \\
    warmup epochs & 10 \\
    training epochs & 100 \\
    augmentation & RandomResizedCrop \\
    \hline
    \end{tabular}
    \label{tab:linear}
\end{table}

\end{document}